\documentclass{article} 
\usepackage{iclr2015,times}
\usepackage{hyperref}
\usepackage{url}

\usepackage{epsfig}
\usepackage{graphicx}
\usepackage{amsmath}
\usepackage{amssymb}

\usepackage{url}
\usepackage[caption=false]{subfig}
\usepackage{mdwlist}
\usepackage[lined, boxed]{algorithm2e}
\usepackage{multirow}
\usepackage{amsthm}
\usepackage{color}

\graphicspath{{figs/}}

\title{Random Forests Can Hash}

\author{Qiang Qiu, Guillermo Sapiro, and Alex Bronstein \\
Duke University and Tel Aviv University\\
{\tt\small  \{qiang.qiu, guillermo.sapiro\}@duke.edu; bron@eng.tau.ac.il}\\
}

\iclrfinalcopy 

\begin{document}

\maketitle

\begin{abstract}
Hash codes are a very efficient data representation needed to be able to cope with the ever growing amounts of data. We introduce a random forest semantic hashing scheme with information-theoretic code aggregation, showing for the first time how random
forest, a technique that together with deep learning have shown spectacular results
in classification, can also be extended to large-scale retrieval.
 Traditional random forest fails to enforce the consistency of hashes generated from each tree for the same class data, i.e., to preserve the underlying similarity, and it also lacks a principled way for code aggregation across trees.
We start with a simple hashing scheme, where independently trained random trees in a forest are acting as hashing functions.
We the propose a subspace model as the splitting function, and show that it enforces the hash consistency in a tree for data from the same class. We also introduce an information-theoretic approach for aggregating codes of individual trees into a single hash code, producing a near-optimal unique hash for each class.
Experiments on large-scale public datasets are presented, showing that the proposed approach
significantly outperforms state-of-the-art hashing methods for retrieval tasks.
\end{abstract}

\section{Introduction}
\label{sec:intr}

In view of the recent huge interest in image classification and object recognition problems and the spectacular success of deep learning and random forests in  these tasks, it seems astonishing that much less efforts are being invested into related, and often more difficult, problems of image content-based retrieval, and, more generally, similarity assessment in large-scale databases. These problems, arising as primitives in many computer vision tasks, are becoming increasingly important in the era of exponentially increasing information.
Semantic and similarity-preserving hashing methods have recently received considerable attentions to address such a need, in part due to their  memory and computational advantage over other representations.
These methods learn to embed data points into a space of binary strings; thus producing compact representations with constant or sub-linear search time.
Such an embedding can be considered as a hashing function on the data, which translates the underlying similarity into the collision probability of the hash or, more generally, into the similarity of the codes under the Hamming metric.
Examples of recent similarity-preserving hashing methods include \cite{LSH}, \cite{KLSH}, \cite{SH}, \cite{sparsehash}, \cite{KSH}, \cite{AGH}, \cite{hash_hd}, and \cite{STH}.

Due to the conceptual similarity between the problems of semantic hashing and that of binary classification, numerous classification techniques have been adapted to the former task.
For example, state-of-the-art supervised hashing techniques like \cite{sparsehash}, \cite{MM-NN}, and \cite{HDML} are based on deep learning methodologies.
Random forest \citep{RF2001, RFBook} is another popular classification techniques.
Random forests have not been so far used to construct semantic hashing schemes.
This is mainly because acting as a hashing function, a random forest fails to preserve the underlying similarity due to the inconsistency of hash codes generated in each tree for the same class data; it also lacks a principled way of aggregating hash codes produced by individual trees into a single longer code.

This work is the first construction of a semantic hashing scheme based on a random forest. We first introduce a  transformation learner model for random forest  enforcing the hash consistency in a tree, thereby preserving similarity. Then, we propose an information-theoretic approach for aggregating hash codes in a forest, encouraging a unique code for each class.
Using challenging large-scale examples,
we demonstrate  significantly more \emph{consistent} and \emph{unique} hashes for data from the same semantic class, when compared to other state-of-the-art hashing schemes.

\section{Forest Hashing}
\label{sec:thm}

Random forest \citep{RF2001, RFBook} is an ensemble of binary \emph{decision trees}. Following the random forest literature,  in this paper, we specify a maximum tree depth $d$ and also avoid post-training operations such as tree pruning. Thus, a tree of depth $d$ consists of $2^d-2$ tree nodes, excluding the root node, indexed in the breadth-first order. During the training, we introduce randomness into the forest through a combination of random set sampling and randomized node optimization, thereby avoiding duplicate trees.  As discussed in \cite{RF2001} and \cite{RFBook}, training each tree with a different randomly selected set decreases the risk of overfitting,  improves the generalization of classification forests, and significantly reduces the training time. When given more than two classes, we \emph{randomly} partition the classes arriving at each binary split node into two categories for node randomness.

A pedagogic hashing scheme is constructed  as follows: Each data point is pushed through a tree until reaching the corresponding leaf node. We simply set `1' for nodes visited, and `0' for the rest.
By ordering those bits in a predefined node order, e.g.,  the breadth-first order, we obtain a $(2^d-2)$-bit sparse hash code, always containing exactly $d-1$ ones.
In a random forest consisting of $M$ trees of the depth $d$, each point is simultaneously pushed through all trees to obtain $M$ $(2^d-2)$-bit hash codes. Both the training and the hashing processes can  be done in parallel to achieve high computational efficiency on modern parallel CPU or GPU hardware.

   In classification, for which the forest was originally designed,  an ensemble posterior is obtained by averaging from a large number of trees, thus boosting the classification accuracy \citep{RF2001}, and  no confident class posteriors are required for individual trees.
However, due to the lack of confident class posteriors for individual trees,
 we obtain highly inconsistent hashes from an individual tree for the same class data. It is also not obvious how to combine hashes from different trees given a target code length.
The inconsistency of the hash codes prevents standard random forest from being directly adopted for hashing, being such codes critical for large-scale retrieval.

To address these problems we first propose a transformation as the learner model for the random forest \citep{forest_qiu, lowrankT}, where each tree enforces consistent codes for similar points. Though a class may not be assigned a unique code in each tree due to limited leaf availability, each class shares code with different classes in different trees due to the underlying node randomness models.

 We further propose an information-theoretic approach to aggregate hashes across trees into a unique code for each data class.
 Consider a random forest consisting of $M$ trees of depth $d$; the hash codes obtained for $N$ training samples are denoted as $\mathcal{B} = \{ \mathbf{B}_i\}_{i=1}^M$, with the $\mathbf{B}_i \in \{0,1\}^{(2^d-2) \times N}$ being the codes generated from the $i$-th tree, henceforth denoted as \emph{code blocks}. Given the target hash code length $L$, our objective is to select $k$ code blocks $\mathbf{B}^*$, $k \le L/(2^d-2)$, maximizing the mutual information between the selected and the remaining codes,
$
\mathbf{B}^* = \arg \max_{\mathbf{B}: |\mathbf{B}|=k } I(\mathbf{B}; \mathcal{B} \backslash \mathbf{B}).
 \label{eqt:infomax1}
$
When the class labels ${C}$ are available for a subset of training samples, semi-supervised aggregation is performed as
$
\mathbf{B}^* = \arg \max_{\mathbf{B}: |\mathbf{B}|=k } I(\mathbf{B}; \mathcal{B} \backslash \mathbf{B}) + \lambda I(\mathbf{B}; {C}).
 \label{eqt:infomax3}
$
The two terms here can be evaluated using different samples to exploit all labeled and unlabeled data. Note that the code aggregation step is only learned once during training, no cost at testing.

\section{Experimental Evaluation}
\label{sec:exp}

\begin{figure*} [t]
\centering
 \subfloat[Six different image queries in CIFAR-10.] {\label{fig:cifar} \includegraphics[angle=0, height=0.25\textwidth, width=.48\textwidth]{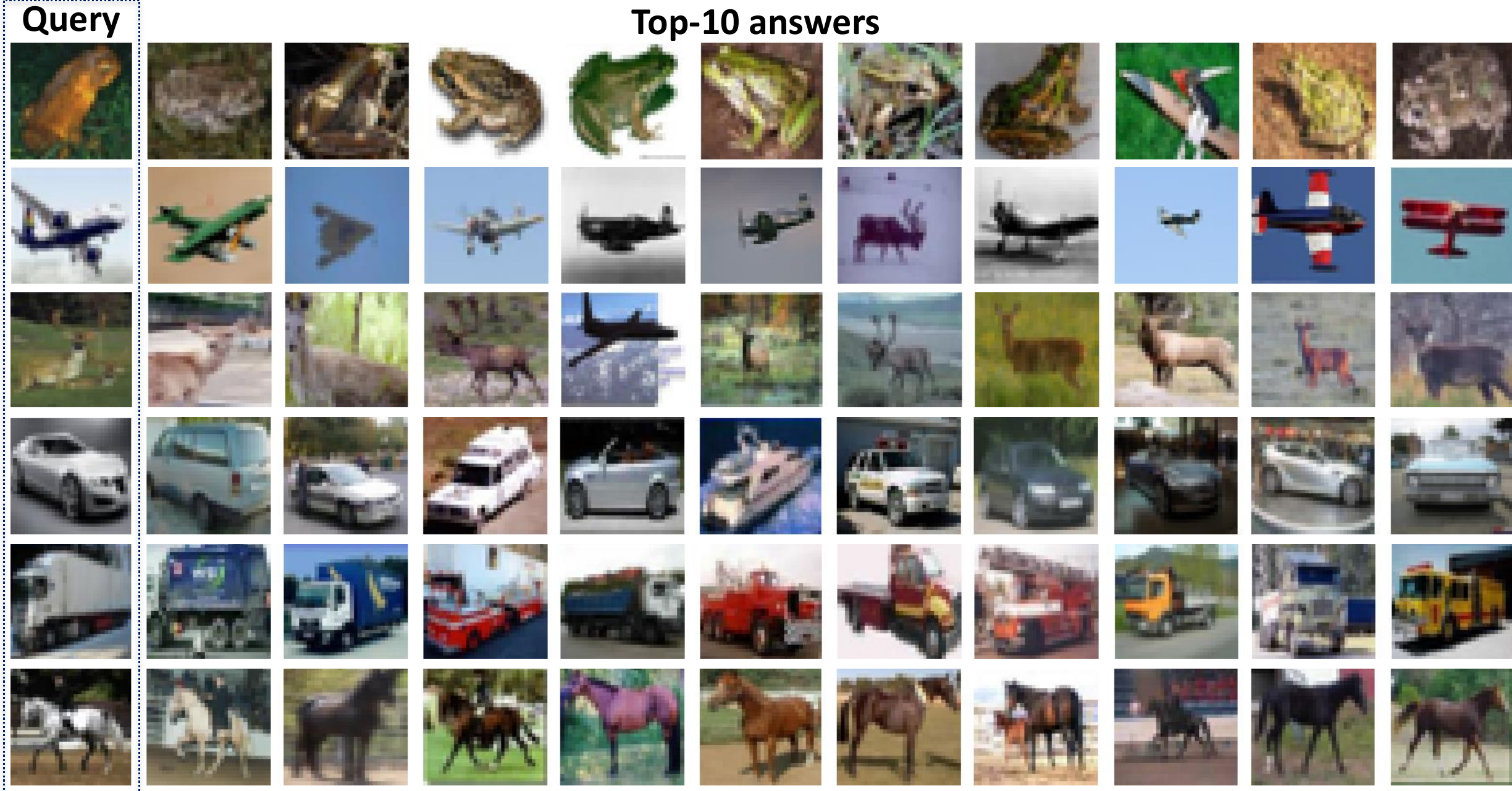} \hspace{0pt}}
 \subfloat[Six different face queries in Pubfig.] {\label{fig:imgquery} \includegraphics[angle=0, height=0.25\textwidth, width=.48\textwidth]{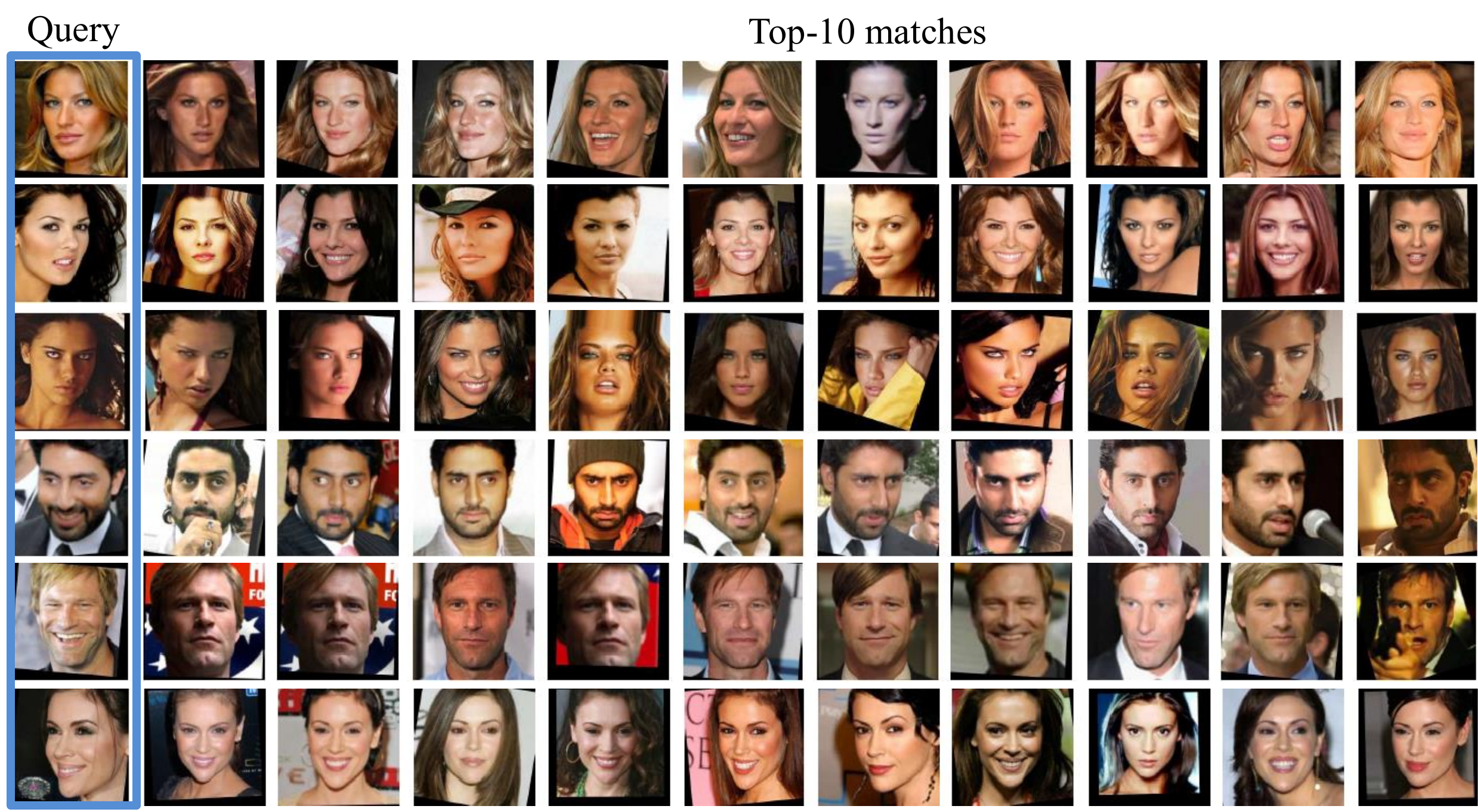}}
\caption{Ten top matches retrieved using the proposed technique.}
\label{fig:toyexample}
\end{figure*}

We present an experimental evaluation of ForestHash on retrieval tasks using standard public benchmarks.
Hashing methods compared include supervised methods FastHash \citep{fasthash},  TSH \citep{tsh}, HDML \citep{HDML}, KSH \citep{KSH},  LDAHash \citep{DH}, and unsupervised methods SH \citep{SH}, KLSH \citep{KLSH}, AGH \citep{AGH}. All software was provided by the authors.

 We adopt the same setup as in \cite{HDML} for the image retrieval experiments on MNIST.
We trained a forest of 64 trees of depth 3. Table~\ref{tab:hdml} summarizes the retrieval performance of various methods at  Hamming radius 0.
Here HDML is a deep learning based hashing method, and FastHash is a booted trees based method.
We denote the proposed method as \emph{ForestHash}.
Due to our subspace-based leaner models, which are known to be robust for small training samples \cite{revlearn}, and our semi-supervised code aggregation that exploits both labeled and unlabeled data, ForestHash significantly outperforms state-of-the-art  methods for reduced training cases.

\begin{table*}[ht]
\centering
	\caption{36-bit retrieval performance (\%) on MNIST (rejection hamming radius 0) using different  training set sizes. Test time is the average binary encoding time in microseconds ($\mathrm{\mu s}$).}
{\small
	\begin{tabular}{l|l|l l|l l|l l}
	\hline
	\hline
 & & \multicolumn{2}{c|}{6,000 samples per class}  & \multicolumn{2}{c|}{ 100 samples per class} & \multicolumn{2}{c}{ 30 samples per class} \\
 	\cline{3-8}
&Test time ($\mathrm{\mu s}$) &Precision & Recall & Precision& Recall  & Precision& Recall \\
	\hline
TSH  & 411 &  86.30   & 3.17    &    74.00  &  5.19  & 56.86  &  3.94        \\
HDML  &10 & {92.94}  &  60.44     &  62.52  &  2.19    & 24.28  &  0.21    \\
FastHash  & 115 & 84.70  &  {76.60}  &      73.32   & 33.04     &      57.08 &  11.77    \\
\textbf{ForestHash} & 17 &  88.81 &   68.54    &   {86.86}  &  {65.72}    & {79.19}  & { 57.93}  \\
\hline
	\hline
	\end{tabular}
}	
	\label{tab:hdml}
\end{table*}

We adopt a challenging setup as in \cite{KSH, sparsehash} for the image retrieval experiments on CIFAR10 \citep{cifar10}.
Table~\ref{tab:cifar10} summarizes the retrieval performance of various methods for Hamming radius 0 and 2.
ForestHash-base is the pedagogic random forest hashing scheme in Section~\ref{sec:thm}, where the decision stump learner model is used and a random subset of trained trees are selected.
It is surprising that this simple pedagogic scheme outperforms all compared supervised methods at radius 0 with orders of magnitude speedup, and the recall is significantly improved with the proposed code aggregation (aggr.). ForestHash using the transformation learner dramatically improves the precision over the pedagogic scheme, significantly outperforms all compared methods at radius 0, and reports comparable precision and significantly higher recall at radius 2.
Figure~\ref{fig:cifar} presents image query examples in the CIFAR-10 dataset.

\begin{table*}[ht]
\centering
	\caption{{36-bit retrieval performance ($\%$) on CIFAR10.
} }
{ \small
	\begin{tabular}{l|l|ll|ll}
	\hline
	\hline
 & & \multicolumn{2}{c|}{radius = 0}  & \multicolumn{2}{c}{radius $\le$ 2}  \\
 	 	\cline{3-6}
 &  Test time ($\mathrm{\mu s}$) & Precision & Recall & Precision& Recall \\
	\hline
AGH1 & 14 &29.59 &   0.25   &    29.79 &   0.71  \\
AGH2 & 20 & 29.45  &  0.32  &  27.70  &  1.07       \\
 	\hline
KSH & 16 &  13.88   & 0.11   &  33.18   & 1.10  \\
FastHash & 94 &14.86 &   0.53  &      36.13  &  2.25   \\
TSH & 206 & 15.80  &  0.23    &   36.99  &  2.47   \\
	\hline
{ForestHash-base}& 0.6&17.24 &   4.37  &    14.86 &   9.92       \\ 
{ForestHash-base (aggr.)}& 0.6 &15.95  &  12.54  &     14.37  &  21.44    \\ 
{\textbf{ForestHash}}& 14 &32.47  &  5.90    &    31.06  &  11.28     \\ 
 \hline
 	\hline
	\end{tabular}
}	
	\label{tab:cifar10}
\end{table*}

Results reported in  Table~\ref{tab:pubfigsmall} refer to an experiment on the Pubfig face dataset \citep{faceattr} , in which we construct the hashing forest using 30 training faces per subject (5,992 faces from 200 subjects ), and  search among their 37,007 unseen faces.
As subspace methods are robust for small training samples problems \citep{revlearn} and extraordinarily effective in representing faces \citep{Wright09}, ForestHash shows significantly higher precision and recall compared to all state-of-the-art methods.
Figure~\ref{fig:imgquery} presents several examples of face queries.

\begin{table}[ht]
\centering
	\caption{{36-bit retrieval performance ($\%$)  on  the Pubfig face dataset (rejection radius 0), 5,992 queries (200 known subjects)  over 37,007 unseen faces of query subjects.
}}
{\small
	\begin{tabular}{l|l|l|l}
	\hline
	\hline
 & Test time ($\mathrm{\mu s}$) & Precision & Recall \\
	\hline
SH & 8 & 9.23  &  0.21    \\
KLSH &  15& 22.09  &  4.05    \\
AGH1& 10 &33.37 &   54.17   \\
AGH2 & 16 & 25.85  &  58.10     \\
\hline
LDAHash &2 &30.05  &  1.28   \\
FastHash& 97  &58.95  &   3.47  \\
TSH & 693& 17.41  &   0.22 \\
\hline
\textbf{ForestHash} & 28& 97.72   &  85.12 \\
 \hline
 	\hline
	\end{tabular}
}	
	\label{tab:pubfigsmall}
\end{table}

\section{Conclusion}
\label{sec:con}

Considering the importance of compact and computationally efficient codes, we introduced a random forest semantic hashing scheme, which, to the best of our knowledge, is the first instance of using random forests for hashing, extending the use of it beyond classification  for large-scale  retrieval.
The proposed scheme consists of a forest with  transformation learners, and an information-theoretic code aggregation scheme.
The proposed framework combines in a fundamental fashion feature learning, random forests, and similarity-preserving hashing, and can be straightforwardly extended to retrieval of incommensurable multi-modal data.
Our method shows exceptional effectiveness in preserving similarity in hashes, and outperforms
state-of-the-art hashing methods in  large-scale retrieval tasks.

\appendix

{
\bibliographystyle{iclr2015}
\bibliography{hash}
}

\end{document}